\documentclass[letterpaper]{article} 
\usepackage[draft]{aaai25}  
\usepackage{times}  
\usepackage{helvet}  
\usepackage{courier}  
\usepackage[hyphens]{url}  
\usepackage{graphicx} 
\usepackage{natbib}  
\usepackage{caption} 
\usepackage{algorithm}
\usepackage{algorithmic}
\usepackage{booktabs}
\usepackage{multirow}
\usepackage{bm}
\usepackage{newfloat}
\usepackage{listings}
\DeclareCaptionStyle{ruled}
{labelfont=normalfont,labelsep=colon,strut=off} 
\floatstyle{ruled}
\newfloat{listing}{tb}{lst}{}
\floatname{listing}{Listing}
\title{DiverseDialogue: A Methodology for Designing Chatbots with Human-Like Diversity}
\author{
    Xiaoyu Lin\textsuperscript{\rm 1},
    Xinkai Yu\textsuperscript{\rm 1},
    Ankit Aich\textsuperscript{\rm 1, 2},
    Salvatore Giorgi\textsuperscript{\rm 2},
    Lyle Ungar\textsuperscript{\rm 1}
}
\affiliations{
    \textsuperscript{\rm 1}University of Pennsylvania\\
    \textsuperscript{\rm 2}National Institutes of Health\\

    \{linxy1, xinkaiyu, ungar\}@seas.upenn.edu,
    \{ankit.aich, sal.giorgi\}@nih.gov 
}

\begin{document}

\maketitle

\begin{abstract}
Large Language Models (LLMs), which simulate human users, are frequently employed to evaluate chatbots in applications such as tutoring and customer service. Effective evaluation necessitates a high degree of human-like diversity within these simulations. In this paper, we demonstrate that conversations generated by GPT-4o mini, when used as simulated human participants, systematically differ from those between actual humans across multiple linguistic features. These features include topic variation, lexical attributes, and both the average behavior and diversity (variance) of the language used. To address these discrepancies, we propose an approach that automatically generates prompts for user simulations by incorporating features derived from real human interactions, such as age, gender, emotional tone, and the topics discussed. We assess our approach using differential language analysis combined with deep linguistic inquiry. Our method of prompt optimization, tailored to target specific linguistic features, shows significant improvements. Specifically, it enhances the human-likeness of LLM chatbot conversations, increasing their linguistic diversity. On average, we observe a 54 percent reduction in the error of average features between human and LLM-generated conversations. This method of constructing chatbot sets with human-like diversity holds great potential for enhancing the evaluation process of user-facing bots.

\end{abstract}

%

\section{Introduction}

Chatbots are used in many scenarios that involve conversations, e.g.: bot friends (Replika, \cite{info:doi/10.2196/16235}), bot teachers (Khanmigo,  \cite{khan2023khanmigo}) and bot therapists (Woebot, \cite{info:doi/10.2196/40242}). These bots need to be evaluated on many dimensions, such as correctness and style  (empathy and formality) \cite{DBLP:conf/emnlp/ErsoyVMM23}. Evaluating a chatbot requires collecting conversations \cite{wang2024survey}. However, creating human-bot conversations is often prohibitively expensive \cite{deriu2020spot}.

\begin{figure}[!tb]
\centering
\includegraphics[width=1.0\linewidth]{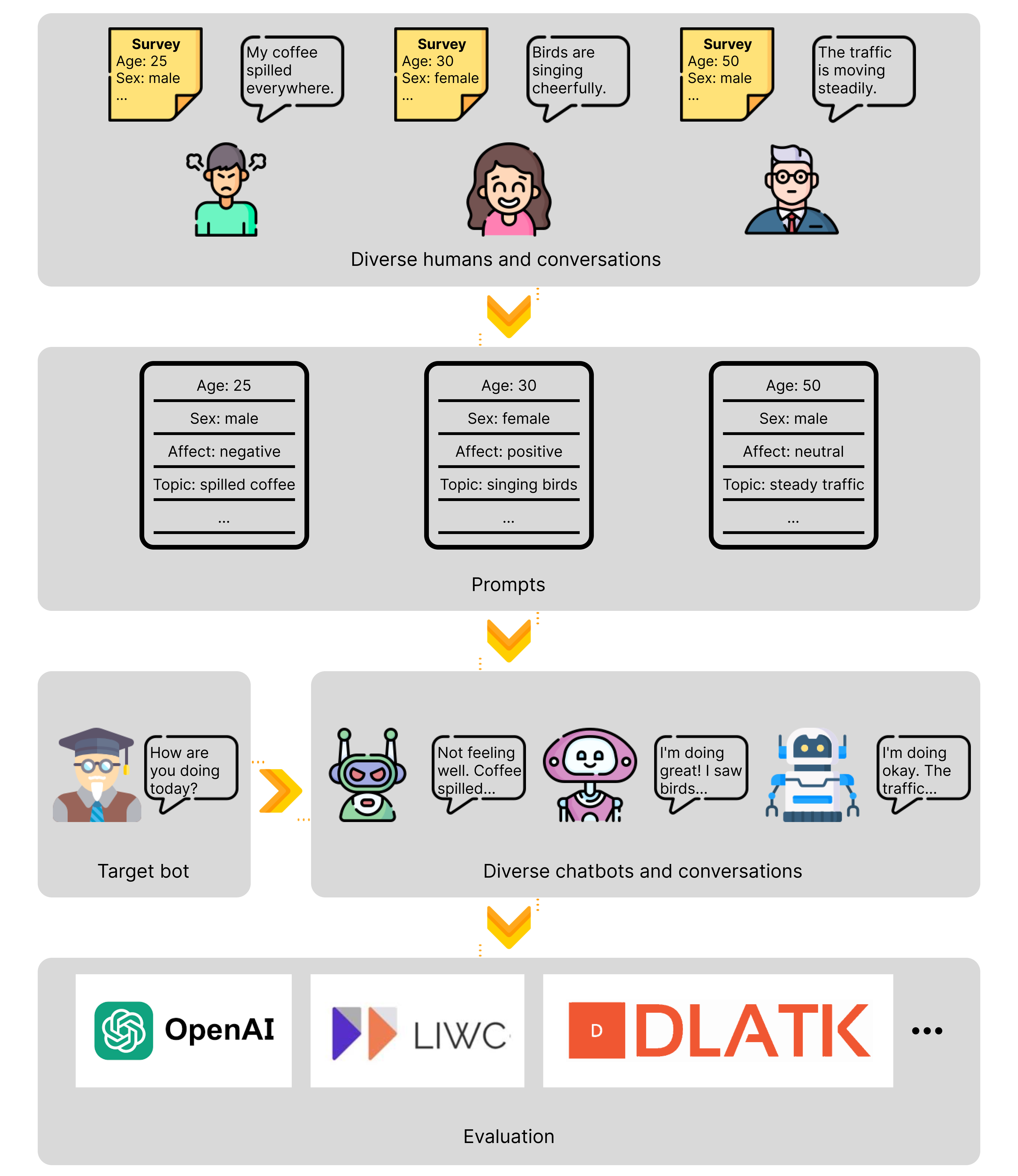}
\caption{DiverseDialogue methodology overview. We use the information (surveys and conversations) about the humans we want to simulate to generate prompts. Survey features (e.g., age, sex, and affect) are obtained directly from surveys, and conversation features (e.g., topic) are summarized from conversations by tools (e.g., GPT-4o mini). These prompts are then used to generate diverse chatbots and conversations. Target bots (e.g. tutors, therapists or friends) then interact with the chatbots that simulate diverse users; the resulting conversations are then evaluated using tools like GPT-4o mini, LIWC, or DLATK.} 
\label{fig:spiritfig}
\end{figure}

An alternative is to have the "target" chatbots that are being evaluated talk to bots simulating users \cite{NEURIPS2019_fc981212}. However, talking to only one "user" bot fails to expose the target chatbot to the wide range of behaviors exhibited by human users. Chatbots are increasingly used in behavioral studies to approximate human subjects, but such human simulations rarely approximate actual populations \cite{moon2024virtualpersonaslanguagemodels}.

In this work, we propose a methodology for generating chatbots that capture human diversity. Because prompt engineering is comparable to fine-tuning \cite{lake2024distributional}, our methodology focuses on constructing effective prompts based on experimental evidence. This makes our algorithm convenient and scalable. Chatbots that follow assigned personas have demonstrated impressive contextual response capabilities in many applications \cite{samuel2024personaagents}. In our DiverseDialog method, we use features such as demographics, personality, and conversation topics to construct chatbot personas for use in simulating real human conversations. 

Figure \ref{fig:spiritfig} shows the overall process. We extract features from a corpus of conversations between different people. We use those to generate prompts that correspond to the different people and their conversations. Then we have the target bot, being evaluated, talk to each of the diverse chatbots. Finally, we evaluate the resulting conversations with GPT-4o mini, psychologically driven lexica, such as Linguistic Inquiry and Word Count (LIWC) \cite{LIWC}, and NLP evaluation packages such as the Differential Language Analysis Toolkit (DLATK) \cite{dlatk}.


Previous research \cite{Peter2023} has highlighted the reduced diversity in standard LLMs when simulating human participants in social science studies. Our work addresses this issue by basing our personas on actual human data. \citeauthor{Argyle_2023} employed data from the American National Election Studies to generate prompts for GPT-3, resulting in simulated responses that reflected the voting behavior and attitudes of various demographic groups. Building on this approach, we extend the strategy to include the extraction of topics, styles, and demographic information, enabling the creation of diverse chatbots that accurately simulate specific real-world populations.

Our methodology, termed DiverseDialog, utilizes an automated algorithm to generate a diverse set of chatbots for the subsequent evaluation of target bots. This approach effectively combines the advantages of the rich diversity inherent in human subjects with the cost-efficiency and convenience of using bots as "test subjects."

Our work introduces a novel approach to the generation and evaluation of a diverse chatbot population. By employing the DiverseDialog approach, we have created a chatbot conversation dataset that aligns with the CANDOR benchmark \cite{reece2023candor}. Additionally, we propose an automated method for evaluating this dataset.

\paragraph{Our specific contributions include:}

\begin{itemize}
\item A novel method for generating chatbots that exhibit human-like diversity.
\item A comprehensive understanding of the features necessary for constructing human-like chatbots, including demographics (age, gender), affect, topics of interest, and style (turn length, formality).
\item The development of an innovative dataset of GPT-4o mini conversations, wherein the language model emulates human interactions based on the pre-existing CANDOR benchmark dataset.
\item A case study analysis comparing the language distribution between humans and chatbots, focusing on demographics, affect, topics of interest, and style.
\end{itemize}
\section{Related Work}
\subsection{Chatbot Simulation of Human Populations }
The simulation of human behavior using chatbots has become an essential practice in many domains, including having AI agents interact and learn from one another within simulated environments \cite{park2023generative}. Similarly, simulated doctors learned how to improve their diagnostic skills in a simulated hospital with simulated patients \cite{li2024agent}. A chatbot for English education can simulate teachers teaching English \cite{lee-etal-2023-peep}. Also, simulated AI subjects behave like human ones, enabling researchers to reproduce classic economic, psycholinguistic, and social psychology experiments \cite{aher2023using}. For example, researchers employ LLMs as substitutes for humans in game experiments \cite{Fan_Chen_Jin_He_2024}. Our goal is to support the automatic generation method of chatbots that simulate populations of people in order make the above tasks faster and more convenient.

\subsection{Diverse Chatbots Generation}
Diversity plays an important role in chatbot generation. Conversational styles of Large Language Models (LLMs) tend to differ systematically from those of humans, both in their average behavior and in their variance \cite{huang2024characterizingdivergences}. Researchers are using in-context
learning to prompt LLMs to generate robust and linguistically diverse output with the goal of simulating the behavior of human interlocutors because such diversity is  useful for evaluating task-oriented dialogues \cite{davidson2023usersimulationlargelanguage}. 

Designers generally try to avoid making chatbots that generate harmful or negative content \cite{hong2023cyclealign}, but when using chatbots as simulated users to evaluate target bots, it is important to simulate real human distributions, including the potentially angry or racist language that users might use. For example, We cannot evaluate teacher bots using only chatbots that simulate good students who like to learn; teacher bots need to be able to handle bad or hostile students. our DiverseDialogue methodology seeks to generate useful distributions of chatbots.

\subsection{LLMs-based Chatbots}
Chatbots are often given specific personas and roles, such as Chinese doctor Zhongjing \cite{Yang_Zhao_Zhu_Zhou_Xu_Jia_Zan_2024}. Three approaches are widely used to build personas: handcrafting, LLM generation, and dataset alignment \cite{wang2024survey}. Manually prompting LLMs is flexible but labor-intensive. LLM generation uses some handcrafted personas as seeds, and then uses LLMs to generate more personas. It is fast and flexible, but does not reflect real human distributions. In dataset alignment, personas are obtained from real-world datasets, which is fast and reflects real-world scenarios. We use a variant of dataset alignment that uses a data set containing both survey data about the people in a population and styles and topics we extract from their conversations. 

\section{Diverse Chatbots Generation}
We propose a two-step method to generate diverse chatbots: diverse information extraction and prompt optimization (including prompt construction, prompt evaluation, and results check). Pseudo-code of the algorithm is given in Algorithm \ref{alg:diverse_agents_generation}.

\begin{algorithm}[tb]
\caption{Diverse Chatbots Generation}
\label{alg:diverse_agents_generation}
\textbf{Input}: Human dialogue dataset $D_h$, Human dialogue prompts $P_h$, Dialogue generation model $M$\\
\textbf{Output}: Diverse dialogue responses $R_{best}$, Prompts $P_{best}$
\begin{algorithmic}[1] 
    \STATE \textbf{Step 1: Diverse Features Extraction}
    \STATE Extract diverse features $F$ from $D_h$

    \STATE \textbf{Step 2: Prompt Optimization}
    \STATE Let $P_{best}=P_h$
    \STATE Generate dialogue responses $R_{best}$ using $M$ with prompts $P_{best}$
    \WHILE{$True$}
        \STATE \textbf{Step 2.1: Prompt Construction}
        \STATE Construct a set of prompts $P$ using diverse features $F$ and the feedback from last iteration (if applicable)

        \STATE \textbf{Step 2.2: Prompt Evaluation}
        \STATE Generate dialogue responses $R$ using $M$ with prompts $P$
        \STATE Apply evaluation tools to both human dialogue $D_h$ and generated dialogue $R$
        
        \STATE \textbf{Step 2.3: Results Check}
        \IF{$R$ performs better than $R_{best}$}
            \STATE Let $R_{best}=R$ and $P_{best}=P$
        \ENDIF
        \IF{Better $R$ can not be found}
            \STATE \textbf{break}
        \ENDIF
    \ENDWHILE
\STATE \textbf{return} $R_{best}$, $P_{best}$
\end{algorithmic}
\end{algorithm}

\subsection{Diverse Features Extraction}
\label{sec:step1}
As emphasized in the introduction, our goal is to generate diverse chatbots that mirror the diversity found in human populations. To achieve this, it is essential to extract and incorporate the diverse features of the individuals we aim to simulate. These features can be categorized into four key dimensions:

\begin{itemize}
\item \textbf{Demographics}: including attributes such as age and gender.
\item \textbf{Affect}: capturing the emotional tone and sentiment.
\item \textbf{Topics of Interest}: focusing on the subjects that engage different groups.
\item \textbf{Style}: encompassing aspects such as turn length and formality in conversation.
\end{itemize}

In this paper, our case study analysis is grounded in the CANDOR dataset. We extract demographic and affective features from survey data, derive topics of interest from dialogue text generated by GPT-4o mini, and manually design conversational style elements.

\subsection{Prompt Optimization}
\label{sec:step2}
\subsubsection{Prompt Construction}
To get the most out of GPT-4o mini in designing diverse chatbots, we build prompts that generate dialogues whose distribution is as close as possible to that of the real people's distribution that is simulated.

Below, we study the relative contribution of different components of the prompt, by changing one attribute at a time. For example, to explore the 'best' prompt for formality, we may have two prompts to compare: 1. Please try to use informal language, the way people talk casually; 2. Please talk like a normal person holding a conversation. The only differences between the two experiments should be these specific prompts.

\subsubsection{Prompt Evaluation}
We evaluate our prompts using a variety of tools, shown in Table \ref{table:promptmodel}.

\begin{table}[]
\centering
\begin{tabular}{@{}ccc@{}}
\toprule
 & Tools & Details\\ 
\midrule
Age & DLATK & Lexica: Age and Gender Lexica\\
Gender & DLATK & Lexica: Age and Gender Lexica\\
Affect & LIWC & Category: Affective processes\\
Topic & DLATK & LDA Topics: 2000 Facebook Topics\\
Formality &  StyLEx & Model: "formal" \\
Length & Python & Programming by hand\\
\bottomrule
\end{tabular}
\caption{Evaluation tools for different aspects of prompts.}
\label{table:promptmodel}
\end{table}

DLATK is an open-source Python package and command-line tool designed for conducting social-scientific language analyses \cite{dlatk}. In our study, we utilize DLATK to evaluate age, gender, and topic features within dialogues. For age and gender, DLATK derives predictive lexica—words and their corresponding weights—using regression and classification models based on word usage in Facebook, blog, and Twitter data, which are annotated with demographic labels. The model achieves a Pearson correlation of r=0.831 for age prediction and a binary gender classification accuracy of 0.919 \cite{sap2014agegender}. Additionally, we employ 2,000 publicly available LDA topics, created using Mallet \cite{Mallet}, within DLATK. These topics are derived from Facebook posts.

LIWC (Linguistic Inquiry and Word Count) is a text analysis application that offers an efficient and effective method for analyzing the emotional, cognitive, and structural components present in both verbal and written speech samples \cite{LIWC}. In this paper, we utilize LIWC 2015. The core of LIWC's text analysis strategy is its set of dictionaries (lexica). Each dictionary entry corresponds to one or more word categories or subdictionaries. For example, the word "cried" falls into five categories: sadness, negative emotion, overall affect, verbs, and past focus. Consequently, when the word "cried" appears in the target text, the scale scores for each of these five subdictionaries are incremented. We use LIWC2015 specifically to evaluate affect, focusing on the category labeled "affective processes."

StyLEx \cite{hayati-etal-2023-stylex} is a joint model for predicting style at both the word and sentence levels. It incorporates a transformer-based encoder, a word-level style predictor, and a sentence-level style predictor. For our purposes, we employ the formal model of StyLEx to assess style in the dialogues.

Turn length is simply calculated by counting (in python) the number of words in each turn. 

\subsubsection{Results Check}
We compute the following features for both the human and chatbot dialogues:

\begin{itemize}
\item Scalar feature: age, gender, affect, formality, length
\item Vector feature: distribution over 2000 topics
\end{itemize}

We then compare a variety of different metrics over these features:
 
\begin{table*}[]
\centering
\begin{tabular}{@{}cccc@{}}
\toprule
Data Type & Average Error & Error of Average & Error of  Dispersion \\ 
\midrule
Scalar& $\frac{1}{n} \sum_{i=1}^{n} \frac{| x_{C_i} - x_{L_i} |}{| x_{C_i} |}$ & $\frac{|\bar{X}_{C} - \bar{X}_{L}|}{|\bar{X}_{C} |}$ & $\frac{|\sigma_{X_{L}}-\sigma_{X_{C}}|}{|\sigma_{X_{C}}|}$ \\
Vector& $\frac{1}{n} \sum_{i=1}^{n} \frac{|| \bm{x_{C_i}} - \bm{x_{L_i}} ||}{|| \bm{x_{C_i}} ||}$ & $\frac{||\bm{\bar{x}_{C}} - \bm{\bar{x}_{L}}||}{||\bm{\bar{x}_{C}} ||}$  & $\frac{||\Sigma_{L} - \Sigma_{C}||}{||\Sigma_{C}||}$ \\

\bottomrule
\end{tabular}
\caption{Speaker-level metrics for evaluating prompts. All normalized by the actual value in the CANDOR dataset. n is the number of speakers. $x_{C_i}$ and $\bm{x_{C_i}}$ are the scalar and vector features of CANDOR, respectively. $\bar{X}_{C}$ and $\bm{\bar{x}_{C}}$ are mean value of $x_{C_i}$ and mean vector of $\bm{x_{C_i}}$, respectively. $\sigma_{X_{C}}$ and $\Sigma_{C}$ are standard deviation of $x_{C_i}$ and covariance matrix of $\bm{x_{C_i}}$, respectively. The mathematical symbol of LLM dialogue is similar, which is denoted as $L$.}
\label{table:metrics}
\end{table*}

Table \ref{table:metrics} presents the metrics employed in this study, offering a comprehensive approach to comparing and evaluating the performance of different dialogues.

For scalar features, the {\it average error} metric represents the mean error between the CANDOR and LLM dialogue features. Each error term is calculated as the absolute difference between two scalar values, normalized by the reference value from CANDOR, thus converting it to a more interpretable scale. The {\it error of average} refers to the relative difference between the mean values of the CANDOR and LLM dialogue features. This metric provides an intuitive measure of overall bias by comparing the average performance of the two dialogues on a global scale. Lastly, the {\it error of dispersion} calculates the normalized difference in feature distributions between the two dialogues, offering insights into the variability and spread of features.

For topic distributions, we use analogous metrics adapted to vector spaces by employing vector norms of differences instead of absolute values. These metrics enable us to quantify how the results of chatbot conversations, generated from different prompts, deviate from the source CANDOR conversations in terms of both their means and variances. This allows for a nuanced assessment of the similarity between the dialogues.  

\section{Experiment}
In this section, we demonstrate our DiverseDialog approach by showing how information can be extracted from a set of human dialogue from CANDOR and how the extracted features can then be used in an interactive prompt optimization process to generate a diverse set of chatbots.

\subsection{Human Conversation Data}
\label{sec:data}
We use the CANDOR Corpus \cite{reece2023candor} as our human conversation data set. CANDOR contains 1650 conversations that strangers had over video chat along with rich metadata information obtained from pre-conversation and post-conversation surveys. The corpus draws on a large and diverse sample of participants, aged 19-66, from all around the United States \cite{reece2022advancing}. Therefore, it is a good data set to build diverse dialogue chatbots. We use the following information from CANDOR.

\begin{itemize}
\item Age: "age" in the survey, a number between 19 and 66;
\item Gender: "sex" in the survey, male or female;
\item Affect: "overall\_affect" in the survey: a number between 1 and 9;
\item Transcription: based on audiophile algorithm (CANDOR dataset provides the transcription processed by this algorithm).
\end{itemize}

We use the 1603 of the 1650 conversations that have no missing values.

\subsection{Human Conversation Data Preprocessing}
\label{sec:datapreprocess}
To construct appropriate prompts, we need to preprocess affect and transcription. Age and gender do not need to be processed; we can use them from the CANDOR survey directly.

\subsubsection{Affect}
We convert numbers into natural language. The scale in the CANDOR survey ranges from 1 to 9, with lower numbers indicating negative sentiments, such as "extremely negative" or "moderately negative", and higher numbers indicating positive sentiments, such as "extremely positive" or "moderately positive". A middle value, like 5, represents a neutral sentiment. This scale translates numerical values into descriptive language to express varying degrees of emotional tone.

\subsubsection{Transcription}
To summarize topics of transcription. We use GPT-4o mini by prompting "Please summarize the topic of the following sentences in 4 words or less:". Because the CANDOR conversations are relatively long, we only input from the 51st to the 70th turns to the GPT-4o mini. We do not use the first 50 turns which mostly contain sentences that start the conversations or debug equipment, like "Hi", "How are you?", "Could you hear me?".

\subsection{Evaluation tools output}
The features computed on the CANDOR and LLM-generated dialogues are mostly used as extracted, with the exception of language-estimated gender, which is discretized to 1 (female) or -1 (male)

\subsection{Prompt Optimization: Formality}
\label{sec:promptoptimization}
Here, we show prompt optimization of formality as an example. Each time a new prompt is tried, we generate 100 conversations, with each conversation consisting of 64 turns.

\subsubsection{Prompt: System Role}
The system role is used to set the behavior and personality of the assistant. It establishes the rules and context within which the assistant operates. This initial message is typically used to convey information that the assistant needs to follow during the conversation, such as style or specific guidelines.

To explore good prompts of formality, we try three system prompts (the rest of the system prompts are the same):
\begin{itemize}
\item A: No prompts about formality;
\item B: "Please try to use informal language, the way people talk casually";
\item C: "Please talk like a normal person holding a conversation".
\end{itemize}

\subsubsection{Prompt: User Role}
In our experiments, each utterance serves as a prompt for another chatbot. The first prompt is an exception: "Talk about whatever you like, just imagine you have met someone at a social event and you're getting to know each other." This prompt is the same prompt that was given to the human participants in the original CANDOR data collection.

\subsubsection{Formality Prompts Evaluation}
We use the error of average to evaluate the prompts. The results are shown in Table \ref{table:formal}. We can see that ”Please try to use informal language, the way people talk casually" is the best formality system prompt because the difference between CANDOR and dialogue generated by prompt B is smallest on average. Hence, we choose it as our final formality prompt.

\begin{table}[]
\centering
\small
\begin{tabular}{@{}cccc@{}}
\toprule
 Prompts & $\bar{X}_{C}$ & $\bar{X}_{L}$ & $\frac{|\bar{X}_{C} - \bar{X}_{L}|}{|\bar{X}_{C} |}$
  \\ \midrule
A   & 0.54 & 0.98 & 0.83\\
B   & 0.54 & 0.82 & 0.52\\
C   & 0.54 & 0.99 & 0.83\\ \bottomrule
\end{tabular}
\caption{Evaluating formality prompts. $\bar{X}_{C}$ and $\bar{X}_{L}$ are turn-level mean values of LIWC2015 informal feature for CANDOR and GPT-4o mini, respectively.}
\label{table:formal}
\end{table}

\subsection{The Performance of the Best Prompts}
In this section, we will show the performance of the best prompts after all of the optimization. The best prompts are:

\begin{itemize}

\item {\em System role:} Imagine that you are a [Gender] and [Age] years old. You feel [Affect] feelings. You should not explicitly say that you have these characteristics, but your conversation should be typical of someone with these characteristics. Please try to use informal language, the way people talk casually. Please say at most one or two sentences per turn.

\item {\em User role:} Talk about whatever you like, just imagine you have met someone at a social event and you're getting to know each other. If it fits the conversation, you should talk about one of the following topics: [Topic].

\end{itemize}

[Age], [Gender], [Affect], and [Topic] are obtained from the feature extraction on the human conversations. Some of the above prompts are fixed across all studies; They do not contribute to the diversity but make dialogues more human-like.

\begin{table}[]
\centering
\normalsize
\begin{tabular}{ccccccccc}
\toprule
& Prompt Level& $\bar{X}_{C}$ & $\bar{X}_{L}$ &  $\frac{|\bar{X}_{C} - \bar{X}_{L}|}{|\bar{X}_{C} |}$  \\ 
\midrule
  \multirow{2}*{\rotatebox{0}{Length}} 
 & No Prompts  & 6.66 & 455.46  & 67.39 \\
 & Best Prompts  & 6.66 & 16.16 & 1.43 \\
 \midrule
  \multirow{2}*{\rotatebox{0}{Formality}} 
 & No Prompts   & 0.54 & 1.00 & 0.86 \\
 & Best Prompts  & 0.54 & 0.82 & 0.52 \\
\bottomrule
\end{tabular}
\caption{DiverseDialogue performance: The features in this table contribute to making the dialogue more human-like. All the computations are at turn level. $n$ is the number of speakers in CANDOR dataset. $x_{C_i}$ is the sentence length or informal level of CANDOR. $\bar{X}_{C}$ is mean value of $x_{C_i}$. $\sigma_{X_{C}}$ is standard deviation of $x_{C_i}$. The mathematical symbol of LLM dialogue is similar, which is denoted as $L$. The closer the error of average value is to zero, the better the performance of the generated dialogue.}
\label{tab:scalar1 turn-level results}
\end{table}

\begin{table*}[]
\centering
\normalsize
\begin{tabular}{ccccccccc}
\toprule
& Prompt Level& $\bar{X}_{C}$ & $\bar{X}_{L}$ & $\sigma_{X_{C}}$& $\sigma_{X_{L}}$ & $\frac{1}{n} \sum_{i=1}^{n} \frac{| x_{C_i} - x_{L_i} |}{| x_{C_i} |}$ & $\frac{|\bar{X}_{C} - \bar{X}_{L}|}{|\bar{X}_{C} |}$ & $\frac{|\sigma_{X_{L}}-\sigma_{X_{C}}|}{|\sigma_{X_{C}}|}$ \\ 
\midrule
\multirow{2}*{\rotatebox{0}{Age}} 
 & No Prompts & 19.62 & 30.70 & 4.00 & 5.14 & 0.65 & 0.56 & 0.29 \\
 & Best Prompts & 19.62 & 28.36 & 4.00 & 3.91 & 0.51 & 0.45 & 0.02 \\
 \midrule
 \multirow{2}*{\rotatebox{0}{Gender}} 
 & No Prompts  & 0.89 & 0.07 & 0.45 & 1.00 & 0.91 & 0.92 & 1.21 \\
 & Best Prompts& 0.89 & 0.88 & 0.45 & 0.48 & 0.20 & 0.02 & 0.07\\
 \midrule
 \multirow{2}*{\rotatebox{0}{Affect}} 
 & No Prompts  & 0.04 & 0.09 & 0.01 & 0.02 & 1.15 & 1.06 & 1.41\\
 & Best Prompts & 0.04 & 0.06 & 0.01 & 0.01 & 0.52 & 0.45 & 0.21\\
\bottomrule
\end{tabular}
\caption{DiverseDialogue performance: The features in this table contribute to enabling the chatbot to exhibit human-like diversity. All the computations are at speaker level. $n$ is the number of speakers in CANDOR dataset. $x_{C_i}$ is the age, gender, or affect of CANDOR. $\bar{X}_{C}$ is mean value of $x_{C_i}$. $\sigma_{X_{C}}$ is standard deviation of $x_{C_i}$. The mathematical symbol of LLM dialogue is similar, which is denoted as $L$. The closer the average error / error of average / error of dispersion value is to zero, the better the performance of the generated dialogue.}
\label{tab:scalar2 speaker-level results}
\end{table*}

\begin{table*}[]
\centering
\normalsize
\begin{tabular}{ccccc}
\toprule
& Prompt Level & $\frac{1}{n} \sum_{i=1}^{n} \frac{|| \bm{x_{C_i}} - \bm{x_{L_i}} ||}{|| \bm{x_{C_i}} ||}$ & $\frac{||\bm{\bar{x}_{C}} - \bm{\bar{x}_{L}}||}{||\bm{\bar{x}_{C}} ||}$  &  $\frac{||\Sigma_{L} - \Sigma_{C}||}{||\Sigma_{C}||}$  \\ 
\midrule
\multirow{2}*{\rotatebox{0}{Topic}} 
 & No Prompts  & 1.45  & 1.23 &  1.10\\
 & Best Prompts  & 1.27  & 1.12 & 1.06\\
\bottomrule
\end{tabular}
\caption{DiverseDialogue performance: The features in this table contribute to ensuring that the chatbots' dialogue topics align more closely with human conversations. All the computations are at speaker level. n is the number of speakers in CANDOR dataset. $\bm{x_{C_i}}$ is the vector representation of the topic of the CANDOR conversations, containing the conditional probability for each topic. $\bm{\bar{x}_{C}}$ is mean vector of $\bm{x_{C_i}}$. $\Sigma_{C}$ is covariance matrix of $\bm{x_{C_i}}$. The mathematical symbol of LLM dialogue is similar, which is denoted as $L$. The closer the error of dispersion value is to zero, the better the performance of the generated dialogue.}
\label{tab:vector speaker-level results}
\end{table*}

In this case study, we evaluated the performance differences between GPT-4o mini and the CANDOR benchmark under various prompting conditions, as illustrated in Table \ref{tab:scalar1 turn-level results}, Table \ref{tab:scalar2 speaker-level results}, and Table \ref{tab:vector speaker-level results}. We observed varying impacts of prompts across different aspects of diversity. The experimental results demonstrate that our DiverseDialogue methodology improves the performance of the GPT-4o mini in terms of human-like diversity.

For the length and formality in Table \ref{tab:scalar1 turn-level results}, the LLM exhibited noticeable bias when no prompts are used. Because these two prompts contribute only to the human-like aspects of the dialogue and do not change with different chatbots, we only use the mean values to evaluate. Specifically, the LLM's turn length is much higher than the CANDOR, and the introduction of prompts greatly reduces this discrepancy (the error of average value decreases from 67.39 to 1.43), indicating that prompts effectively regulate the LLM's output length. The turn of the original LLM dialogues is too long, and after our method correction, the chatbots' dialogue is more human-like in length. Similarly, the original LLM chatbots speak relatively formally. With our approach, the chatbots speak more like humans, speaking naturally and casually.

By analyzing the age, gender, and affect in Table \ref{tab:scalar2 speaker-level results}, we find that our method simulates diverse populations very well. For the age, the LLM's average value without prompts was notably higher than the CANDOR benchmark (30.70 vs. 19.62) and exhibited a larger standard deviation, indicating the distribution of LLM chatbots is far from the CANDOR people. However, upon introducing optimal prompts, the LLM's average age decreased to 28.36, bringing it closer to the CANDOR benchmark, while the standard deviation was reduced from 5.14 to 3.91. The reduction in the error of dispersion (from 0.29 to 0.02) further confirms the importance of our method.

In the gender evaluation, the effect of our methodology is even more pronounced. Without prompts, the LLM's average gender was markedly off from the CANDOR benchmark (0.07 vs. 0.89), with a standard deviation much higher than that of CANDOR. This indicates that the LLM's initial gender distribution is about half male and half female, but most of people are female in CANDOR. After introducing the best prompts, the LLM's average gender nearly matched the CANDOR benchmark (0.88 vs. 0.89). The error of average dropped dramatically from 0.92 to 0.02 and the error of dispersion is very close to 0, demonstrating that our method can simulate the age distribution of real people very well.

Moreover, in the affect evaluation, prompts also show a positive impact. Without prompts, the LLM's dialogue has higher mean and standard deviation values compared to the CANDOR benchmark, resulting in a relatively high average error. However, the use of prompts narrows these differences, with the average error and error of average reduced by half and the error of dispersion dropped to one-seventh of its original value. This further underscores the importance of DiverseDialogue methodology in controlling the chatbots' affect distribution.

For the topic of interest in Table \ref{tab:vector speaker-level results}, we further validated the effectiveness of prompts. Without prompts, the vectors generated by the LLM exhibited differences from the CANDOR benchmark. However, with the use of optimal prompts, this difference was reduced. This is reflected in the reduction of both the average vector difference and the covariance matrix difference, indicating that our methodology can control the topics LLM chatbots are interested in. This is an important part of chatbot diversity.

We can clearly see that different features have different levels of performance improvement. The errors of length and gender are down nearly 100 percent. However, the topic has only been slightly improved. Compared to GPT-4o mini conversations, CANDOR has a relatively large number of turns and topics. We only extract one topic per CANDOR conversation for prompting LLM. Even so, our method reduces error by about ten percent. This proves that our approach is very promising and there is room for improvement.

The experimental results indicate that DiverseDialogue methodology greatly improves the performance of the LLM chatbots across multiple aspects and metrics. These findings suggest that DiverseDialogue methodology is an effective means of aligning LLM chatbots' diversity to the real population and making them human-like.

\section{Discussion}
\label{sec:discuss}

In this paper, we examine the capacity of large language models to generate human-like diversity in turn-based conversations. Our findings highlight three key areas of discussion.

\paragraph{The Importance of Human-Like Diversity in Chatbot Evaluation}
As outlined in the introduction, chatbots are increasingly employed across a wide range of applications, including companionship, therapy, and training. These domains are inherently human-facing, making accurate representation of human diversity crucial. Prior research by \citet{positive_outcomes_human_style} demonstrates that chatbot interactions characterized by person-centered communication—emphasizing empathy, concern, and tailored communication styles—yield more positive outcomes. In this study, we demonstrate that off-the-shelf chatbots, such as GPT, do not inherently adjust for diversity. We also present methods that can be employed to address this shortcoming, offering a more robust framework for chatbot development.

\paragraph{Quantifiable Evaluation Metrics for Chatbots}
Quantitatively evaluating chatbot performance is inherently challenging due to the broad range of domains they serve, with user satisfaction often being the primary metric of success. In this paper, we seek to mitigate some of these challenges by employing a comparative analysis. Instead of merely assessing chatbot performance in isolation, we benchmark it against a human gold standard. This approach enables us to delve deeper into the limitations of chatbots and observe changes in diverse dialogue generation through differential language analysis and examination of lexical features.

\paragraph{ Considerations in Simulating Human Behavior}
Simulating human behavior presents ethical challenges, particularly concerning authentic representation of diversity versus the risk of perpetuating stereotypes. Chatbots have, at times, exhibited racially biased and stereotypical language when attempting to emulate the communication styles of specific demographic groups \citep{aich2024vernacularibarelyknow}. 

\section{Conclusion}
In this work, we identified major discrepancies between conversations generated by GPT-4o mini based user simulations and those occurring between real human users. These discrepancies were evident across several linguistic dimensions, including formality, turn length, etc. It underscores the limitations of current LLM-based user simulations in achieving human-like diversity.

To address the limitations, we developed a method that constructs prompts for user simulations by extracting features from actual human conversations, such as age, gender, affect, and discussed topics. This approach enhances the human likeness and diversity of LLM-generated chatbots.

Our experimental results demonstrate that simulations using these feature-informed prompts lead to more realistic and varied interactions, closely mirroring the diversity observed in real human conversations. On average, we reduce the error of average between human and LLM conversations by 54 percent, and nearly 100 percent for some features. Also, the error of dispersion is reduced by 70 percent on average. This improvement has the potential to enhance the robustness of chatbot evaluations, particularly in applications such as tutoring and customer service, where understanding and replicating human diversity is crucial.

Future work could explore further refinements to the prompt generation process, including the incorporation of additional human features or more complex conversational dynamics. Additionally, applying our methodology to other LLMs or expanding it to different domains may reveal further insights into the best practices for creating truly human-like simulated users.

\section{Limitations}
While the proposed DiverseDialogue methodology demonstrates large improvements in generating chatbots with human-like diversity, there are several limitations to consider. First, the methodology was primarily validated using the CANDOR dataset, which, although comprehensive, limits the generalizability of the findings. Testing on multiple datasets across various domains could provide a more robust evaluation of the approach. Additionally, the complexity of the multi-step process, which includes feature extraction, prompt optimization, and iterative evaluation, may introduce challenges in reproducibility and scalability. The computational cost associated with these processes is not thoroughly discussed, raising concerns about the practicality of applying this methodology to larger datasets or more complex scenarios. Furthermore, the evaluation relies heavily on statistical comparisons of dialogue features, which, while informative, may not fully capture the subtleties of human-like diversity. Incorporating qualitative analyses or human evaluations could offer a more nuanced understanding of the dialogues' effectiveness. 

\section{Ethical Statement}

The DiverseDialogue methodology presented in this paper has the potential to greatly enhance the evaluation and development of chatbot systems by simulating a more diverse range of human-like interactions. This can lead to more robust and empathetic AI systems, particularly in domains such as education, mental health, and customer service, where understanding and replicating the diversity of human behavior is critical. By improving the realism of these simulations, our work can contribute to the development of AI systems that are better equipped to handle a wide range of user needs and behaviors, ultimately leading to more inclusive and effective technologies.

However, the ability to simulate diverse human behaviors also comes with potential ethical concerns. Our methodology includes the simulation of potentially negative or harmful behaviors, such as hostility or prejudice, which, if not carefully managed, could result in unintended consequences or misuse. There is a risk that such simulations could reinforce harmful stereotypes or be exploited to create manipulative or unethical AI systems. It is crucial that developers employing our methodology implement safeguards to ensure that the generated content is used responsibly and that the ethical implications are carefully considered throughout the development process.

While our work has the potential to contribute positively to society by creating more adaptive and inclusive AI systems.

\bibliography{aaai25}

\begin{thebibliography}{30}
\providecommand{\natexlab}[1]{#1}

\bibitem[{Aher, Arriaga, and Kalai(2023)}]{aher2023using}
Aher, G.~V.; Arriaga, R.~I.; and Kalai, A.~T. 2023.
\newblock Using large language models to simulate multiple humans and replicate human subject studies.
\newblock In \emph{International Conference on Machine Learning}, 337--371. PMLR.

\bibitem[{Aich et~al.(2024)Aich, Liu, Giorgi, Isman, Ungar, and Curtis}]{aich2024vernacularibarelyknow}
Aich, A.; Liu, T.; Giorgi, S.; Isman, K.; Ungar, L.; and Curtis, B. 2024.
\newblock Vernacular? I Barely Know Her: Challenges with Style Control and Stereotyping.
\newblock arXiv:2406.12679.

\bibitem[{Argyle et~al.(2023)Argyle, Busby, Fulda, Gubler, Rytting, and Wingate}]{Argyle_2023}
Argyle, L.~P.; Busby, E.~C.; Fulda, N.; Gubler, J.~R.; Rytting, C.; and Wingate, D. 2023.
\newblock Out of One, Many: Using Language Models to Simulate Human Samples.
\newblock \emph{Political Analysis}, 31(3): 337–351.

\bibitem[{Davidson et~al.(2023)Davidson, Romeo, Shu, Gung, Gupta, Mansour, and Zhang}]{davidson2023usersimulationlargelanguage}
Davidson, S.; Romeo, S.; Shu, R.; Gung, J.; Gupta, A.; Mansour, S.; and Zhang, Y. 2023.
\newblock User Simulation with Large Language Models for Evaluating Task-Oriented Dialogue.
\newblock arXiv:2309.13233.

\bibitem[{Deriu et~al.(2020)Deriu, Tuggener, von Däniken, Campos, Rodrigo, Belkacem, Soroa, Agirre, and Cieliebak}]{deriu2020spot}
Deriu, J.; Tuggener, D.; von Däniken, P.; Campos, J.~A.; Rodrigo, A.; Belkacem, T.; Soroa, A.; Agirre, E.; and Cieliebak, M. 2020.
\newblock Spot The Bot: A Robust and Efficient Framework for the Evaluation of Conversational Dialogue Systems.
\newblock arXiv:2010.02140.

\bibitem[{Ersoy et~al.(2023)Ersoy, Vizcarra, Mayeesha, and Muller}]{DBLP:conf/emnlp/ErsoyVMM23}
Ersoy, A.; Vizcarra, G.; Mayeesha, T.~T.; and Muller, B. 2023.
\newblock In What Languages are Generative Language Models the Most Formal? Analyzing Formality Distribution across Languages.
\newblock In Bouamor, H.; Pino, J.; and Bali, K., eds., \emph{Findings of the Association for Computational Linguistics: {EMNLP} 2023, Singapore, December 6-10, 2023}, 2650--2666. Association for Computational Linguistics.

\bibitem[{Fan et~al.(2024)Fan, Chen, Jin, and He}]{Fan_Chen_Jin_He_2024}
Fan, C.; Chen, J.; Jin, Y.; and He, H. 2024.
\newblock Can Large Language Models Serve as Rational Players in Game Theory? A Systematic Analysis.
\newblock \emph{Proceedings of the AAAI Conference on Artificial Intelligence}, 38(16): 17960--17967.

\bibitem[{Ghandeharioun et~al.(2019)Ghandeharioun, Shen, Jaques, Ferguson, Jones, Lapedriza, and Picard}]{NEURIPS2019_fc981212}
Ghandeharioun, A.; Shen, J.~H.; Jaques, N.; Ferguson, C.; Jones, N.; Lapedriza, A.; and Picard, R. 2019.
\newblock Approximating Interactive Human Evaluation with Self-Play for Open-Domain Dialog Systems.
\newblock In Wallach, H.; Larochelle, H.; Beygelzimer, A.; d\textquotesingle Alch\'{e}-Buc, F.; Fox, E.; and Garnett, R., eds., \emph{Advances in Neural Information Processing Systems}, volume~32. Curran Associates, Inc.

\bibitem[{Hayati et~al.(2023)Hayati, Park, Rajagopal, Ungar, and Kang}]{hayati-etal-2023-stylex}
Hayati, S.~A.; Park, K.; Rajagopal, D.; Ungar, L.; and Kang, D. 2023.
\newblock {S}ty{LE}x: Explaining Style Using Human Lexical Annotations.
\newblock In Vlachos, A.; and Augenstein, I., eds., \emph{Proceedings of the 17th Conference of the European Chapter of the Association for Computational Linguistics}, 2843--2856. Dubrovnik, Croatia: Association for Computational Linguistics.

\bibitem[{Heyn et~al.(2023)Heyn, Torp~Løkkeberg, Ellington, Dulmen, and Eide}]{positive_outcomes_human_style}
Heyn, L.; Torp~Løkkeberg, S.; Ellington, L.; Dulmen, A.; and Eide, H. 2023.
\newblock Understanding the role of positive emotions in healthcare communication – A realist review.
\newblock \emph{Nursing Open}, 10.

\bibitem[{Hong et~al.(2023)Hong, Tu, Chen, Gao, Zhang, and Yan}]{hong2023cyclealign}
Hong, J.; Tu, Q.; Chen, C.; Gao, X.; Zhang, J.; and Yan, R. 2023.
\newblock CycleAlign: Iterative Distillation from Black-box LLM to White-box Models for Better Human Alignment.
\newblock arXiv:2310.16271.

\bibitem[{Huang et~al.(2024)Huang, Rijn, Sucholutsky, Marjieh, and Jacoby}]{huang2024characterizingdivergences}
Huang, D.-M.; Rijn, P.~V.; Sucholutsky, I.; Marjieh, R.; and Jacoby, N. 2024.
\newblock Characterizing Similarities and Divergences in Conversational Tones in Humans and LLMs by Sampling with People.
\newblock arXiv:2406.04278.

\bibitem[{{Khan Academy}(2024)}]{khan2023khanmigo}
{Khan Academy}. 2024.
\newblock Khanmigo.
\newblock Accessed: 2024-05-22.

\bibitem[{Lake, Choi, and Durrett(2024)}]{lake2024distributional}
Lake, T.; Choi, E.; and Durrett, G. 2024.
\newblock From Distributional to Overton Pluralism: Investigating Large Language Model Alignment.
\newblock arXiv:2406.17692.

\bibitem[{Lee et~al.(2023)Lee, Jang, Park, Lee, Seo, Moon, Eo, Lee, Yahya, and Lim}]{lee-etal-2023-peep}
Lee, S.; Jang, Y.; Park, C.; Lee, J.; Seo, J.; Moon, H.; Eo, S.; Lee, S.; Yahya, B.; and Lim, H. 2023.
\newblock {PEEP}-Talk: A Situational Dialogue-based Chatbot for {E}nglish Education.
\newblock In Bollegala, D.; Huang, R.; and Ritter, A., eds., \emph{Proceedings of the 61st Annual Meeting of the Association for Computational Linguistics (Volume 3: System Demonstrations)}, 190--207. Toronto, Canada: Association for Computational Linguistics.

\bibitem[{Li et~al.(2024)Li, Wang, Zhang, Li, Lai, Kang, Ma, and Liu}]{li2024agent}
Li, J.; Wang, S.; Zhang, M.; Li, W.; Lai, Y.; Kang, X.; Ma, W.; and Liu, Y. 2024.
\newblock Agent Hospital: A Simulacrum of Hospital with Evolvable Medical Agents.
\newblock arXiv:2405.02957.

\bibitem[{McCallum(2002)}]{Mallet}
McCallum, A. 2002.
\newblock Mallet: A machine learning for languagetoolkit.
\newblock \emph{http://mallet.cs.umass.edu}.

\bibitem[{Moon et~al.(2024)Moon, Abdulhai, Kang, Suh, Soedarmadji, Behar, and Chan}]{moon2024virtualpersonaslanguagemodels}
Moon, S.; Abdulhai, M.; Kang, M.; Suh, J.; Soedarmadji, W.; Behar, E.~K.; and Chan, D.~M. 2024.
\newblock Virtual Personas for Language Models via an Anthology of Backstories.
\newblock arXiv:2407.06576.

\bibitem[{Nicol et~al.(2022)Nicol, Wang, Graham, Dodd, and Garbutt}]{info:doi/10.2196/40242}
Nicol, G.; Wang, R.; Graham, S.; Dodd, S.; and Garbutt, J. 2022.
\newblock Chatbot-Delivered Cognitive Behavioral Therapy in Adolescents With Depression and Anxiety During the COVID-19 Pandemic: Feasibility and Acceptability Study.
\newblock \emph{JMIR Form Res}, 6(11): e40242.

\bibitem[{Park et~al.(2023)Park, O'Brien, Cai, Morris, Liang, and Bernstein}]{park2023generative}
Park, J.~S.; O'Brien, J.; Cai, C.~J.; Morris, M.~R.; Liang, P.; and Bernstein, M.~S. 2023.
\newblock Generative agents: Interactive simulacra of human behavior.
\newblock In \emph{Proceedings of the 36th Annual ACM Symposium on User Interface Software and Technology}, 1--22.

\bibitem[{Park, Schoenegger, and Zhu(2024)}]{Peter2023}
Park, P.~S.; Schoenegger, P.; and Zhu, C. 2024.
\newblock Diminished diversity-of-thought in a standard large language model.
\newblock \emph{Behavior Research Methods}.

\bibitem[{Pennebaker et~al.(2015)Pennebaker, Boyd, Jordan, and Blackburn}]{LIWC}
Pennebaker, J.; Boyd, R.; Jordan, K.; and Blackburn, K. 2015.
\newblock \emph{The development and psychometric properties of LIWC2015}.
\newblock University of Texas at Austin.

\bibitem[{Reece et~al.(2022)Reece, Cooney, Bull, Chung, Dawson, Fitzpatrick, Glazer, Knox, Liebscher, and Marin}]{reece2022advancing}
Reece, A.; Cooney, G.; Bull, P.; Chung, C.; Dawson, B.; Fitzpatrick, C.; Glazer, T.; Knox, D.; Liebscher, A.; and Marin, S. 2022.
\newblock Advancing an Interdisciplinary Science of Conversation: Insights from a Large Multimodal Corpus of Human Speech.
\newblock arXiv:2203.00674.

\bibitem[{Reece et~al.(2023)Reece, Cooney, Bull, Chung, Dawson, Fitzpatrick, Glazer, Knox, Liebscher, and Marin}]{reece2023candor}
Reece, A.; Cooney, G.; Bull, P.; Chung, C.; Dawson, B.; Fitzpatrick, C.; Glazer, T.; Knox, D.; Liebscher, A.; and Marin, S. 2023.
\newblock The CANDOR corpus: Insights from a large multimodal dataset of naturalistic conversation.
\newblock \emph{Science advances}, 9(13): eadf3197.

\bibitem[{Samuel et~al.(2024)Samuel, Zou, Zhou, Chaudhari, Kalyan, Rajpurohit, Deshpande, Narasimhan, and Murahari}]{samuel2024personaagents}
Samuel, V.; Zou, H.~P.; Zhou, Y.; Chaudhari, S.; Kalyan, A.; Rajpurohit, T.; Deshpande, A.; Narasimhan, K.; and Murahari, V. 2024.
\newblock PersonaGym: Evaluating Persona Agents and LLMs.
\newblock arXiv:2407.18416.

\bibitem[{Sap et~al.(2014)Sap, Park, Eichstaedt, Kern, Stillwell, Kosinski, Ungar, and Schwartz}]{sap2014agegender}
Sap, M.; Park, G.; Eichstaedt, J.; Kern, M.; Stillwell, D.; Kosinski, M.; Ungar, L.; and Schwartz, H.~A. 2014.
\newblock Developing age and gender predictive lexica over social media.
\newblock In \emph{Proceedings of the 2014 conference on empirical methods in natural language processing (EMNLP)}, 1146--1151.

\bibitem[{Schwartz et~al.(2017)Schwartz, Giorgi, Sap, Crutchley, Ungar, and Eichstaedt}]{dlatk}
Schwartz, H.~A.; Giorgi, S.; Sap, M.; Crutchley, P.; Ungar, L.; and Eichstaedt, J. 2017.
\newblock {DLATK}: Differential Language Analysis {T}ool{K}it.
\newblock In Specia, L.; Post, M.; and Paul, M., eds., \emph{Proceedings of the 2017 Conference on Empirical Methods in Natural Language Processing: System Demonstrations}, 55--60. Copenhagen, Denmark: Association for Computational Linguistics.

\bibitem[{Ta et~al.(2020)Ta, Griffith, Boatfield, Wang, Civitello, Bader, DeCero, and Loggarakis}]{info:doi/10.2196/16235}
Ta, V.; Griffith, C.; Boatfield, C.; Wang, X.; Civitello, M.; Bader, H.; DeCero, E.; and Loggarakis, A. 2020.
\newblock User Experiences of Social Support From Companion Chatbots in Everyday Contexts: Thematic Analysis.
\newblock \emph{J Med Internet Res}, 22(3): e16235.

\bibitem[{Wang et~al.(2024)Wang, Ma, Feng, Zhang, Yang, Zhang, Chen, Tang, Chen, Lin et~al.}]{wang2024survey}
Wang, L.; Ma, C.; Feng, X.; Zhang, Z.; Yang, H.; Zhang, J.; Chen, Z.; Tang, J.; Chen, X.; Lin, Y.; et~al. 2024.
\newblock A survey on large language model based autonomous agents.
\newblock \emph{Frontiers of Computer Science}, 18(6): 186345.

\bibitem[{Yang et~al.(2024)Yang, Zhao, Zhu, Zhou, Xu, Jia, and Zan}]{Yang_Zhao_Zhu_Zhou_Xu_Jia_Zan_2024}
Yang, S.; Zhao, H.; Zhu, S.; Zhou, G.; Xu, H.; Jia, Y.; and Zan, H. 2024.
\newblock Zhongjing: Enhancing the Chinese Medical Capabilities of Large Language Model through Expert Feedback and Real-World Multi-Turn Dialogue.
\newblock \emph{Proceedings of the AAAI Conference on Artificial Intelligence}, 38(17): 19368--19376.

\end{thebibliography}

\end{document}